\documentclass[letterpaper, 10 pt, conference]{ieeeconf}
\IEEEoverridecommandlockouts
\usepackage{cite}
\usepackage{amsmath,amssymb,amsfonts}
\usepackage{algorithmic}
\usepackage{graphicx}
\usepackage{textcomp}
\usepackage{xcolor}
\usepackage{booktabs}
\usepackage{color,soul}
\usepackage{multirow}
\usepackage{colortbl}
\usepackage[export]{adjustbox}
\usepackage{tikz,graphics,color,fullpage,float,epsf,caption,subcaption}
\usepackage[section]{placeins}

\definecolor{mycolor}{HTML}{FFEDA0}

\def\BibTeX{{\rm B\kern-.05em{\sc i\kern-.025em b}\kern-.08em
    T\kern-.1667em\lower.7ex\hbox{E}\kern-.125emX}}

\title{\LARGE \bf
FLAGRED - Fuzzy Logic-based Algorithm Generalizing Risk Estimation for Drones
}

\author{Samuel Hovington$^{1}$, Louis Petit$^{2}$, Sophie Stratford$^{1}$, Philippe Hamelin$^{3}$, \\Alexis Lussier-Desbiens$^{2}$ and François Ferland$^{1}$
\thanks{
This research was funded by Alliance grant number 2601-2600-703 and CRIAQ between Université de Sherbrooke, Hydro-Québec and DroneVolt.
}
\thanks{
$^{1}$Samuel Hovington, Sophie Stratford and François Ferland are with the Department of Electrical and Computer Engineering, Université de Sherbrooke, Québec, Canada.
{\tt\small [samuel.hovington, sophie.stratford, francois.ferland]@usherbrooke.ca}
}%
\thanks{$^{2}$Louis Petit and Alexis Lussier Desbiens are with the Department of Mechanical Engineering, Université de Sherbrooke, Québec, Canada
{\tt\small [louis.petit, alexis.lussier.desbiens]@usherbrooke.ca}
}%
\thanks{$^{3}$Philippe Hamelin is with Hydro-Quebec Research Institute, Varennes, Québec, Canada.
{\tt\small hamelin.philippe@hydroquebec.com}}%
}

\begin{document}

\maketitle
\thispagestyle{empty}
\pagestyle{empty}

\begin{abstract}

Accurately estimating risk in real-time is essential for ensuring the safety and efficiency of many applications involving autonomous robot systems. This paper presents a novel, generalizable algorithm for the real-time estimation of risks created by external disturbances on multirotors. Unlike conventional approaches, our method requires no additional sensors, accurate drone models, or large datasets. It employs motor command data in a fuzzy logic system, overcoming barriers to real-world implementation. Inherently adaptable, it utilizes fundamental drone characteristics, making it applicable to diverse drone models. The efficiency of the algorithm has been confirmed through comprehensive real-world testing on various platforms. It proficiently discerned between high and low-risk scenarios resulting from diverse wind disturbances and varying thrust-to-weight ratios. The algorithm surpassed the widely-recognized ArduCopter wind estimation algorithm in performance and demonstrated its capability to promptly detect brief gusts.
\end{abstract}


\section{Introduction}

Unmanned aerial vehicles (UAVs) are becoming increasingly important for various applications such as inspection\cite{leclerc2023, Petit2022}, surveillance \cite{Shakhatreh2019}, and UAV-assisted networks \cite{ALZAHRANI2020102706, Cheng2018, Mozaffari2019}. 
However, ensuring the safety of these vehicles is of utmost importance to prevent accidents and damage to property. Several approaches have been developed to prevent drone crashes, but given the multitude of factors that can jeopardize the UAV's safety (e.g. hardware failure, communication problem and especially external disturbance), the challenge of accurately assessing the risk level persists. Harsh weather conditions including wind gust is one of the main component that determines drone flyability \cite{Gao2021}. Some applications also require the use of dynamic payload whose effects cannot be eliminated \cite{LaVigne2021, LaVigne2022}. According to Drury et al., in order to achieve effective situational awareness (SA), it is crucial to provide the operator with relevant information regarding the drone's status, the weather conditions surrounding the UAV, and the level of reliability of the UAV\cite{Drury2006}. 

Several approaches have attempted to identify faults produced by failed motors, inaccurate/drifting sensors or communication problems. However, to the authors' knowledge, there is no approach that specifically aims at identifying the risk related to external disturbances. Furthermore, each approach is either platform-dependent, since it either uses a specific numerical model or training dataset, or requires modification of the UAV by adding external sensors. 


Some disturbances 
have a minimal impact on the UAV's performance and can be deemed acceptable. On the other hand, dangerous disturbances can cause actuator saturation, which may lead to instability and could ultimately result in a crash \cite{Liu2011}. The ability to avoid saturation can be measured by the maneuvering margin it possesses. One of the key indicators of this capacity is the motor's margin of maneuver, which is the difference between the motor commands and their saturation point. A larger difference indicates a greater control and response to external disturbances, while a smaller one indicates limited responsiveness.


The significant contributions of this paper lie in its novel approach to determine the level of risk caused by external disturbances impacting multirotor drones in real time, without the necessity for additional sensors, precise drone models, or extensive datasets. Our method accommodates all kinds of external disturbances, making it more general than the existing literature. Its inherent versatility ensures compatibility with any multirotor drone, thereby expanding its scope and practical utility. Furthermore , the proposed framework is modular, allowing it to integrate various aspects such as battery status, sensor signals, and distance measurements, among others. These distinctive features combined enable a more robust, flexible, and comprehensive risk estimation tool for drone applications.

This paper is structured as follows. Section \ref{sec:related_work} presents the latest literature in failure detection and the use of fuzzy logic in the context of drone operation. Section \ref{sec:methodology} details the fuzzy logic estimator, the learning process of the rules and the risk accumulator. Section \ref{sec:experimental_results} presents real life experiments validating the approach. Finally, this paper concludes with future work and potential improvements.
\section{Related Work}
\label{sec:related_work}



The straightforward approach to assess external disturbances is to measure it directly.
Scicluna et al. \cite{Scicluna2023} validated the feasibility of wind measurement using onboard external sensors, demonstrating a measurement accuracy within a margin of 1 m/s. Simon et al. \cite{FlowDrone2022} validated this using a novel flow sensor based on micro-electro-mechanical systems (MEMS) hot-wire technology to precisely measure the wind. However, Abichandani et al. \cite{Abichandani2020} conducted a comparative analysis of various techniques for wind measurement using flow sensors and anemometers. They highlighted the complexities encountered when integrating external sensors on multirotor drones, as opposed to fixed-wing UAVs.

Model-based methods leverage the connections between measurements and predicted states to identify potential malfunctions.
These methods typically involve state estimation with a Kalman filter \cite{Park2017, Zhong2018, Hajiyev2020, Nejati2021, Demitrit2017}. Unfortunately, Kalman filtering necessitates a mathematical representation of the system, which can be challenging for custom drones. 
%
%
Simma et al. \cite{Simma2020} demonstrated that accurate and real-time wind estimation is achievable without employing external sensors. However, their method requires a detailed drone model and relies on experimentally defined parameters like static thrust and drag coefficient.
%
%
In their work, Varigonda et al. \cite{Varigonda2020} implemented an approach using real-time flight parameters to estimate risk, triggering an alert to the pilot whenever the pre-set threshold values were exceeded. While this method circumvents the need for a complex drone model, it still relies on fixed, platform-dependent thresholds. Furthermore, it cannot forecast failures since risk assessment and failure detection only occur during an event.

%
%

Data-driven approaches, while potentially less demanding than model-based apporaches, require training datasets tailored explicitly for the aircraft in question.
%
In their comparative study on fault detection and identification (FDI) methodologies, Freeman et al. \cite{Freeman2013} analyzed the relative effectiveness of model-based and data-driven approaches applied to a fixed-wing aircraft. The investigation confirmed the capability of both methods to accurately detect and identify simple fault scenarios, including aileron malfunction.
%
Brulin et al. \cite{BRULIN2022} employed deep learning techniques to evaluate faults within the propulsion system of an UAV. However, the scope of their work was limited 
to simulated offline data.
Furthermore, their focus remained strictly on internal malfunctions such as Electronic Speed Control (ESC), motor, or propeller defects. A primary drawback of the data-driven approach is the frequent need to expose the UAV to failure conditions, potentially resulting in a crash. However, this issue can be mitigated by understanding the UAV's standard operating state and only flagging an anomaly when residuals exceed noise measurements \cite{Wang2019}.

\subsection{Fuzzy logic}
Defining risk is a complex task because it is not a clear-cut value that can be distinguished by a single limit. The safety or riskiness of a situation often falls on a spectrum rather than being an absolute. Herein, fuzzy logic can offer substantial benefits.
Fuzzy logic excels in generating reliable estimations, especially when the available data is imprecise or insufficient \cite{ZADEH2008}. It is capable of modeling complex, non-linear relations between various variables, like wind speed and drone stability, without the need for high-end computational power.
Fuzzy logic is a model-free approach with the ability to provide effective predictions \cite{Mendel1995}. 
%
In contrast to methods like machine learning, which typically require large datasets, fuzzy logic can be developed with expert insights \cite{Wang1992}. This becomes especially beneficial when data is sparse.
Additionally, fuzzy logic provides more transparency and simplicity than many other methods. The decision-making rules it employs are generally straightforward, making it easier to understand its functionality \cite{Setnes1998}.





Zhang et al. \cite{Zhang2018} introduced a novel Fault Detection and Diagnosis (FDD) methodology, harnessing the capabilities of Fuzzy Logic and Neural Networks (NN) to discern and identify defective sensors. Their approach incorporated the utilization of a Kalman filter to calculate signal residuals, which were subsequently used as inputs to a NN, the neurons of which were characterized as membership functions. The strategy proved successful in diagnosing instances of sensor failure. However, the authors confined their consideration to biases within sensor measurements. 

Thus, employing thrust data within a fuzzy system could be a solution to bypass the difficulties of using an external sensor, developing an accurate model, or collecting specific datasets.

\section{Methodology}
\label{sec:methodology}
\begin{figure}[tbh]
    \centering
    \includegraphics[width=0.5\textwidth]{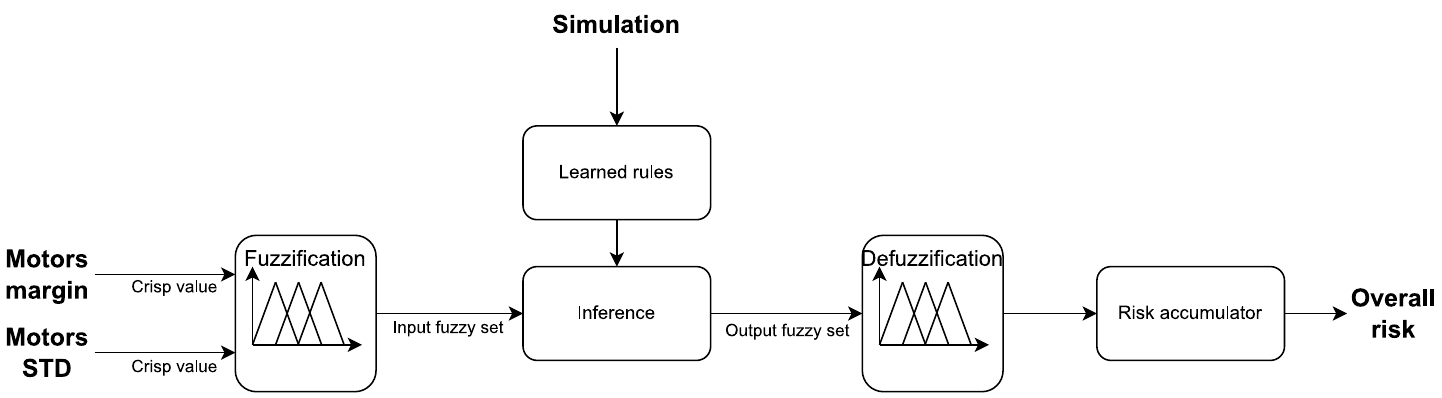}
    \caption{Fuzzy logic overall risk estimator}
    \label{fig:High_level_diagram}
\end{figure}

Our proposed system is shown in Fig. \ref{fig:High_level_diagram} and comprises three principal components: the fuzzy logic estimator, the acquisition of inference rules via simulations, and the accumulation of instantaneous risks.

\subsection{Fuzzy-logic estimator}



\begin{figure*}
  \begin{subfigure}[t]{0.31\textwidth}
    \centering
    \includegraphics[trim={0 0 0 1.4cm},clip,width=\textwidth]{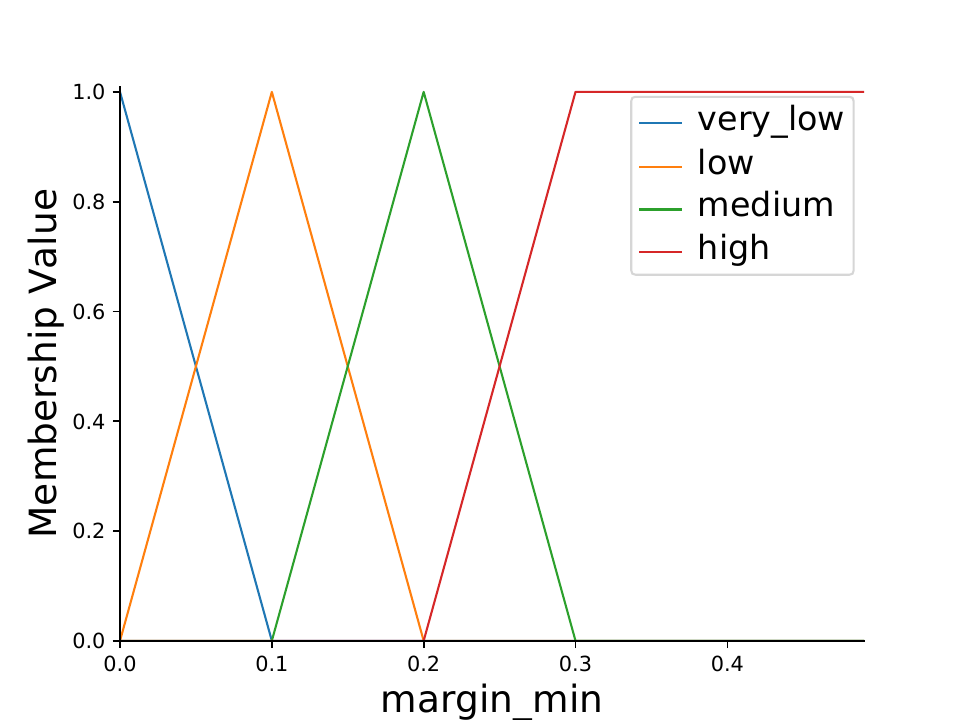}
    \caption{Margin mean input membership functions}
    \label{fig:margin_min_mf}
  \end{subfigure}
  \hfill   
  \begin{subfigure}[t]{0.31\textwidth}
    \centering
    \includegraphics[trim={0 0 0 1.4cm},clip,width=\textwidth]{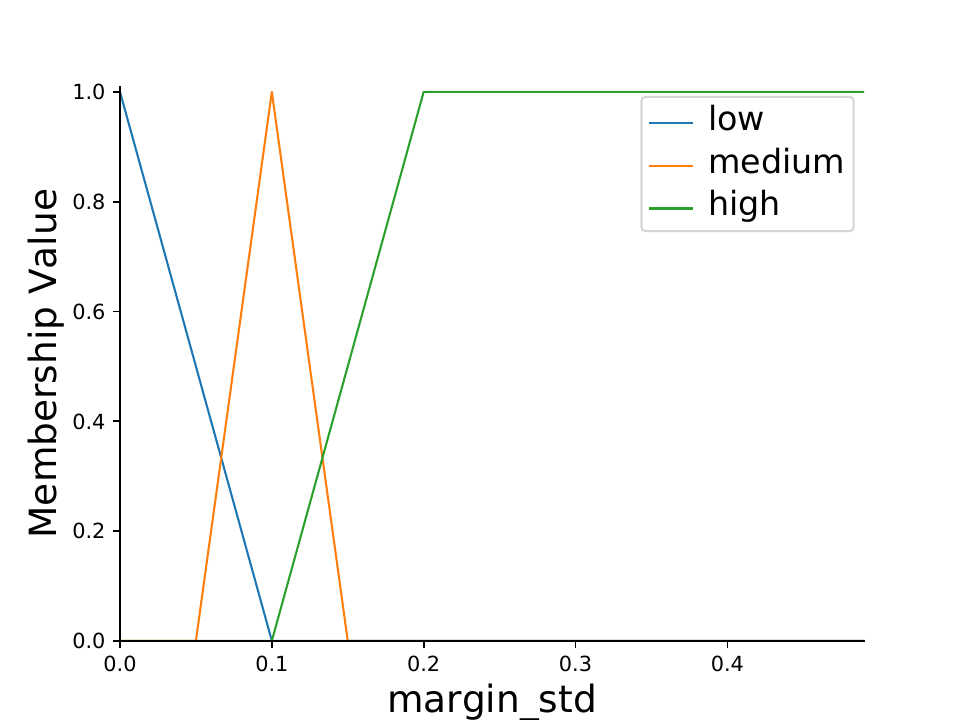}
    \caption{Margin STD input membership functions}
    \label{fig:margin_std_mf}
  \end{subfigure}
  \hfill  
  \centering
  \begin{subfigure}[t]{0.31\textwidth}
    \centering
    \includegraphics[trim={0 0 0 1.4cm},clip,width=\textwidth]{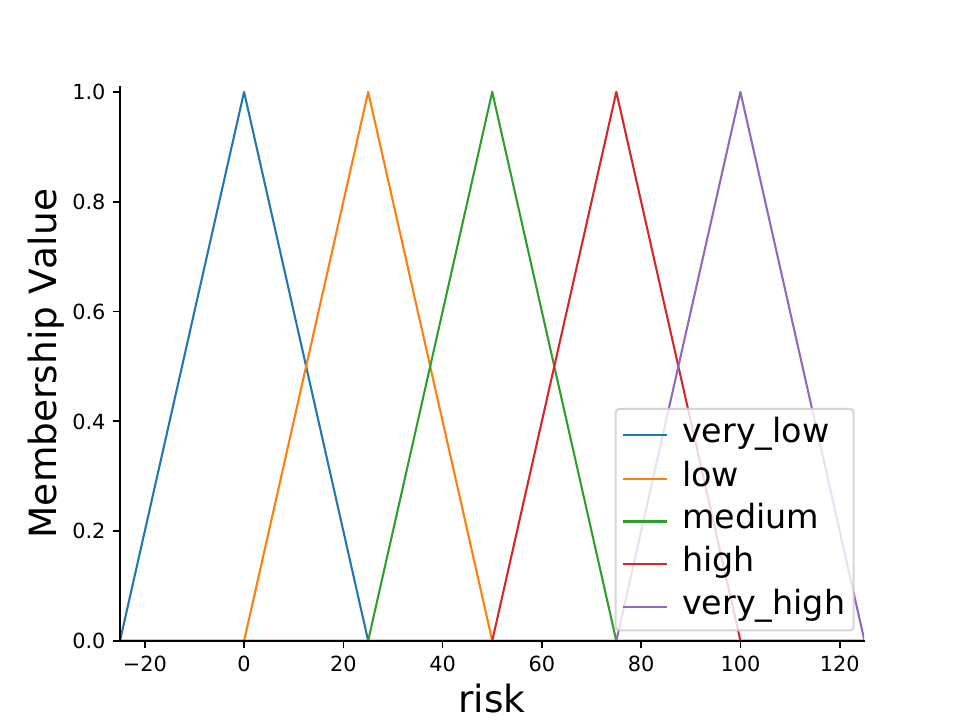}
    \caption{Risk output membership functions}
    \label{fig:risk_mf}
  \end{subfigure}
  \caption{Membership functions used in the Fuzzy logic risk estimator} \label{fig:membership_function}
\end{figure*}

\begin{figure}
    \centering
    \includegraphics[width=0.5\textwidth]{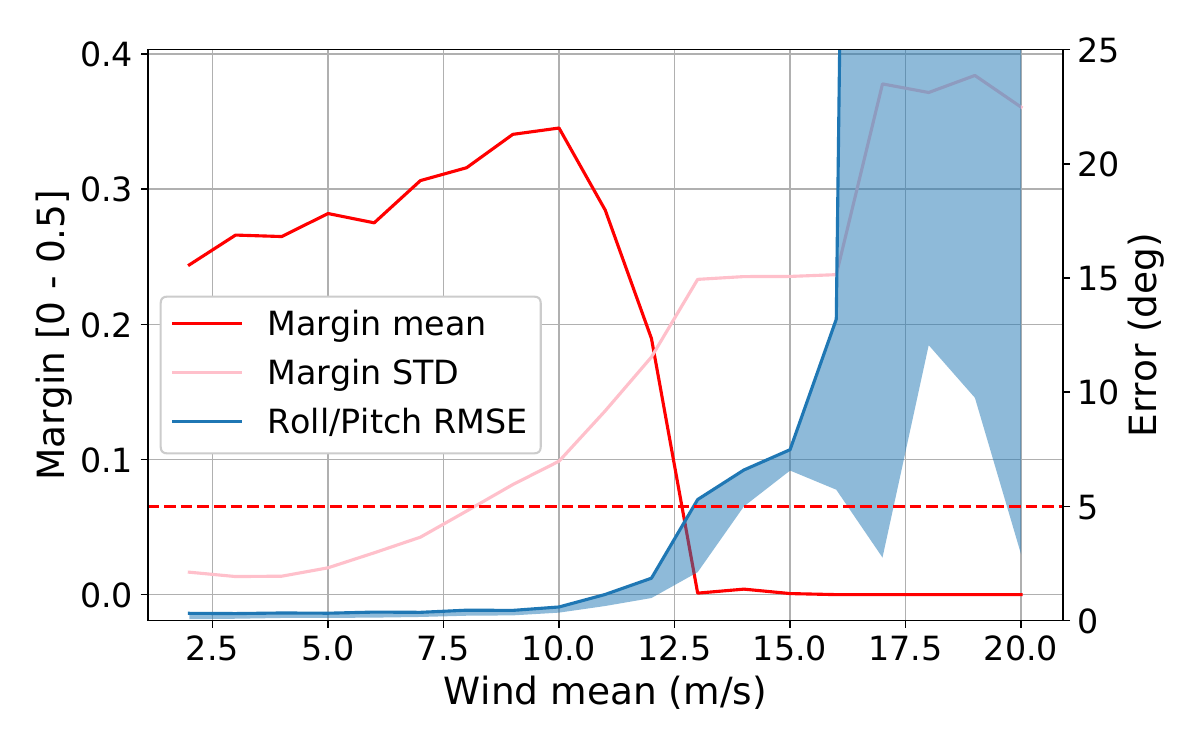}
    \caption{Simulation results to highlight the relation between motor margin and attitude error.}
    \label{fig:simulation_results}
\end{figure}

We emitted the hypothesis that a drone's motor margin is correlated to its ability to withstand external disturbances. We defined this ability by the RMSE in the attitude (roll and pitch) presented in Eq. \ref{eq:error}, with $\phi_{des}$, $\phi$, $\theta_{des}$, and $\theta$ representing the desired and actual roll and pitch angles, respectively. We used Gazebo for physics simulation and Ardupilot Software In The Loop (SITL) for drone control. We simulated 350+ scenarios with wind averages between 0-20 $m/s$ and variances from 0-40 $m^{2}/{s^{2}}$. These ranges were determined empirically, as any wind speed exceeding 18 m/s consistently caused the drone to crash. 
Fig. \ref{fig:simulation_results} presents the results of our simulations. It shows that a drop in margin coincides with an increase in error.

\begin{equation}
    \label{eq:error}
    RMSE = \sqrt{(\phi_{des} - \phi)^{2} - (\theta_{des}-\theta)^{2}}
\end{equation}

To increase modularity and applicability across various types of multirotors, a generalized representation of the drone's margin is needed. This is achieved by using the average and standard deviation of margins during a flight, which are independent of the number of motors on the platform. Most multirotors have Electronic Speed Controllers (ESCs) that convert commands between 1000 and 2000 to motor voltage. Given ESCs' parameters for adjusting saturation to prevent non-linear extremes, we define the motor's normalized command \(c_{\text{norm},n}\) and motor's margin \(m_n\) in equations \ref{eq:c_norm} and \ref{eq:min_distance}. Here, \(n \in [0, N]\), where \(N\) is the number of motors, and \(t_{\text{low}}\) and \(t_{\text{high}}\) denote lower and upper saturation limits, respectively.

\begin{equation}
    \label{eq:c_norm}
   c_{norm,n} = \frac{c_n - t_{low}}{\sqrt{t_{high}^{2} - t_{low}^{2}}}
\end{equation}

\begin{equation} 
    \label{eq:min_distance}
    m_n=min(c_{norm,n}, (1-c_{norm,n}))
\end{equation}



Due to normalization, the margin ranges from 0 to 0.5, establishing the universe of discourse for the mean and standard deviation inputs of our fuzzy logic system. As shown in Fig. \ref{fig:membership_function}, the upper limit of both inputs is considered HIGH, with the residual range divided equally among two or three membership functions. For the output, we have spread the universe across five distinct categories, spanning from VERY LOW to VERY HIGH.


\subsection{Inference rules learning process}


To optimize the fuzzy logic estimator, we need to set inference rules that convert the input parameters (margin mean and standard deviation) into a risk indicator. While expert knowledge is usually crucial for this, defining exact rules for the margin-risk relationship is difficult. Therefore, we have chosen to simulate a drone in various wind conditions, using attitude error as our risk estimation method. We chose 5 degree RMSE as a threshold for defining a critical situation \cite{Gendron2023}.

To derive rules from this dataset, we calculated the mean and standard deviation of margin for each test, pairing these values with the corresponding error, as demonstrated in table \ref{tab:raw_data}. Each data pair was then fuzzified using the previously shown membership functions and assigned with a degree corresponding to the highest degree of ownership. Consequently, we define the degree of a rule as the product of all degrees of ownership for inputs and output. For example, the value of $Margin\_min$, $Margin\_std$ and $Risk$ for the first data pair of Table \ref{tab:raw_data} are respectively $\mu(VERY\_LOW) = 0.920312$,  $\mu(HIGH) = 1.0$  and $\mu(HIGH) = 0.627163$. It results in the following rule: IF Margin\_min is VERY\_LOW and Margin\_std is HIGH, THEN Risk is HIGH with a degree of 0.577186 ($0.920312\times1.0\times0.627163$).

The ensuing step involves identifying all rules possessing identical inputs and retaining only those with the highest rule degree. This procedure culminates in the final set of learned functions, as listed in Table \ref{tab:final_rules}.

    \begin{table}[tbh]
        \centering
        \begin{tabular}{@{}llll@{}}
        \toprule
        \#  & Margin\_min & Margin STD & Risk (\%) \\ \midrule
        0   & 0.007969    & 0.226310   & 65.679063 \\
        2   & 0.070469    & 0.308401   & 100.0     \\
        ... & ...         & ...        & ...       \\
        383 & 0.064687    & 0.153689   & 46.015181 \\
        384 & 0.351094    & 0.092301   & 4.798570  \\ \bottomrule
        \end{tabular}
        \caption{Raw data pairs from the simulation test}
        \label{tab:raw_data}
    \end{table}

    
    \begin{table}
        \centering
        \begin{tabular}{@{}llll@{}}
            \toprule
            Margin   & Margin STD & Risk      & \begin{tabular}[c]{@{}l@{}}Confidence\\ degree\end{tabular} \\ \midrule
            High     & High       & Very low  & 0.23                                                        \\
            High     & Low        & Very low  & 0.88                                                        \\
            High     & Medium     & Very low  & 0.76                                                        \\
            Medium   & High       & Very low  & 0.63                                                        \\
            Medium   & Low        & Very low  & 0.41                                                        \\
            Medium   & Medium     & Low       & 0.73                                                        \\
            Low      & High       & Very high & 0.84                                                        \\
            Low      & Medium     & Medium    & 0.64                                                        \\
            Very low & High       & Very high & 1.0                                                         \\
            Very low & Medium     & Very high & 0.35                                                        \\ \bottomrule
        \end{tabular}
        \caption{Rules learned and their respective confidence degree}
        \label{tab:final_rules}
    \end{table}

\begin{figure}[tbh]
    \centering
    \includegraphics[trim={0 0 0 1.25cm},clip,width=0.5\textwidth]{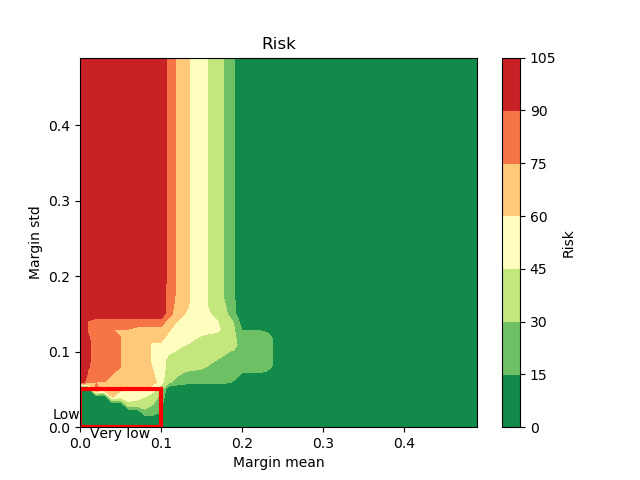}
    \caption{Top-down view of the decision map with highlighted extrapolation zone.}
    \label{fig:decision_map_total}
\end{figure}

The described learning methodology is constrained to producing rules for conditions that are present within the dataset. The dataset was sourced from authentic simulations, leading to some areas being underrepresented. This accounts for the confidence level of 0.23 for the first rule listed in table \ref{tab:final_rules}. Certain areas even remain unexplored. This is evident in the bottom-left corner, highlighted in red, of Fig. \ref{fig:decision_map_total}, where no rules apply to this particular region characterized by a very low margin mean and low standard deviation. To address this limitation, we incorporated a spatial interpolation approach to bridge these gaps and generate a comprehensive decision map, covering all potential input ranges.

\subsection{Elevated risk accumulator}


The fundamental principle of the accumulator is the computation of potential elevated risk in the forthcoming time-step, as informed by the mean and standard deviation of past risk estimations. This process involves the computation of the Probability Density Function (PDF) of risk measurements following a Gaussian distribution, and the use of the Cumulative Distribution Function (CDF) to evaluate the likelihood of a high-risk scenario in the subsequent time-step.\\


Let $R$ be the sequence of previous risk values and let $\mu$ and $\sigma$ be the average and standard deviation of $R$. Because the motor's command follow a normal distribution, we can assume that the PDF of $R$ also follows a normal distribution with parameters $\mu$ and $\sigma$.\\

To calculate $p(r_{high})$, we need to calculate the probability that $r$ exceeds a certain threshold $x$. This can be done with the cumulative distribution function (CDF) following Eq. \ref{eq:p(r_high)} which calculates the probability that the distribution $R$ will take a value less than or equal to $x$.

\begin{equation}
    \label{eq:p(r_high)}
    p(r_{high}) = p(r >  x) = 1-p(r \le  x)
\end{equation}



In practice, the CDF of $R$ can be calculated by transforming it to a standard normal distribution using Eq. \ref{eq:transformation} and then using Eq. \ref{eq:erf} to expressed the CDF in terms of the error function (Eq. \ref{eq:erf2}). 
\begin{equation} 
    \label{eq:transformation}
    Z = \frac{X - \mu}{\sigma}
\end{equation}

\begin{equation}
    \label{eq:erf2}
    \operatorname{erf}(x) = \frac{2}{\sqrt{\pi}} \int_{0}^{x} e^{-t^2} dt
\end{equation}

\begin{equation}
    \label{eq:erf}
    \Phi(x) = p(r \leq x) = \frac{1}{2}\left[1 + \mathrm{erf}\left(\frac{x}{\sqrt{2}}\right)\right]
\end{equation}

We can then use this probability to adjust the rate of risk accumulation. We can reduce the risk accumulation rate when the probability of high risk in the next timestep is low and increase the risk accumulation rate when the probability of high risk is high. We can apply the same logic with the probability of a low risk. This can be achieved by multiplying the risk accumulation or risk reduction rate by the probability of high or low risk, as shown in Eq. \ref{eq:accumulator}:

\begin{equation}
    \label{eq:accumulator}
    \Delta R = k_{i} * p(r_{high}) -k_{d}*p(r_{low}),
\end{equation}

where $\Delta R$ is the change in risk value, $k_i$ and $k_d$ are constants that respectively determines the magnitude of the risk accumulation and risk reduction rate, and $p(r_{high})$ and $p(r_{low})$ are respectively the probability of high and low risk in the next timestep.

\section{Experimental results}
\label{sec:experimental_results}
\begin{figure}
    \centering
    \includegraphics[width=0.4\textwidth]{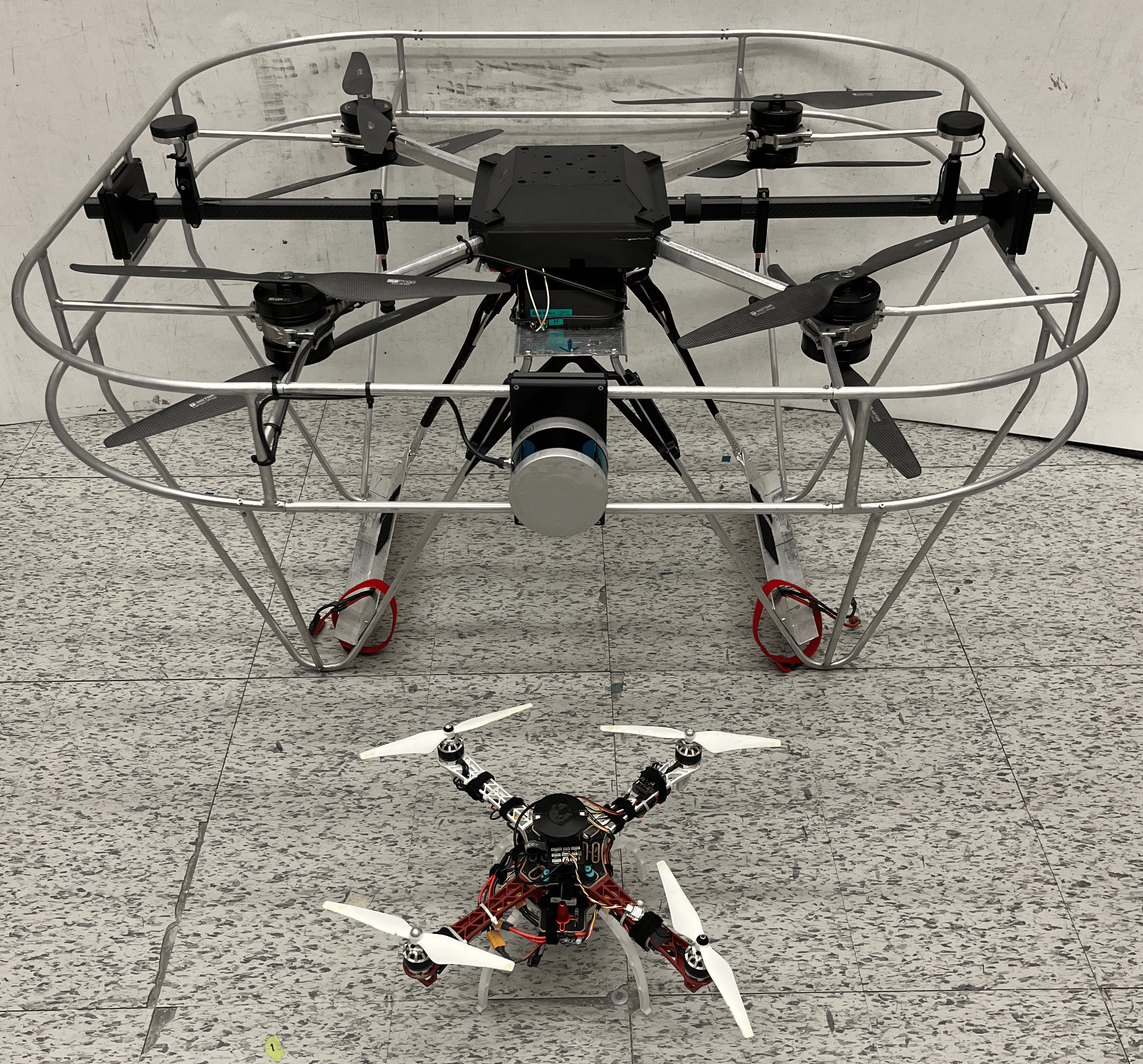}
        \caption{NADILE (Upper) and F450 (Lower)}
        \label{fig:NADILE}
\end{figure}

To validate the feasibility and versatility of our method, we performed experiments in real-world environments using two different platforms. The first is a model of the custom drone created by DroneVolt in partnership with Hydro-Québec Research Institute (IREQ) \cite{Linedrone}. This drone prototype, developed for the Autonomous Navigation of Drones and Interventions on Power Lines (French acronym - NADILE), features an OCTO QUAD X8 setup (see Fig. \ref{fig:NADILE}). The second is the popular DJI F450 quadrotor equipped with the open-source Pixhawk controller.

We opted for the F450 not only due to its popularity and availability but also because its configuration differs significantly from that of NADILE. The F450 is light weight and highly responsive while IREQ's custom drone is slow to react, weighing over 20kg and measuring more that 1.1 meters wide.


\begin{figure}
    \centering
    \includegraphics[width=0.45\textwidth]{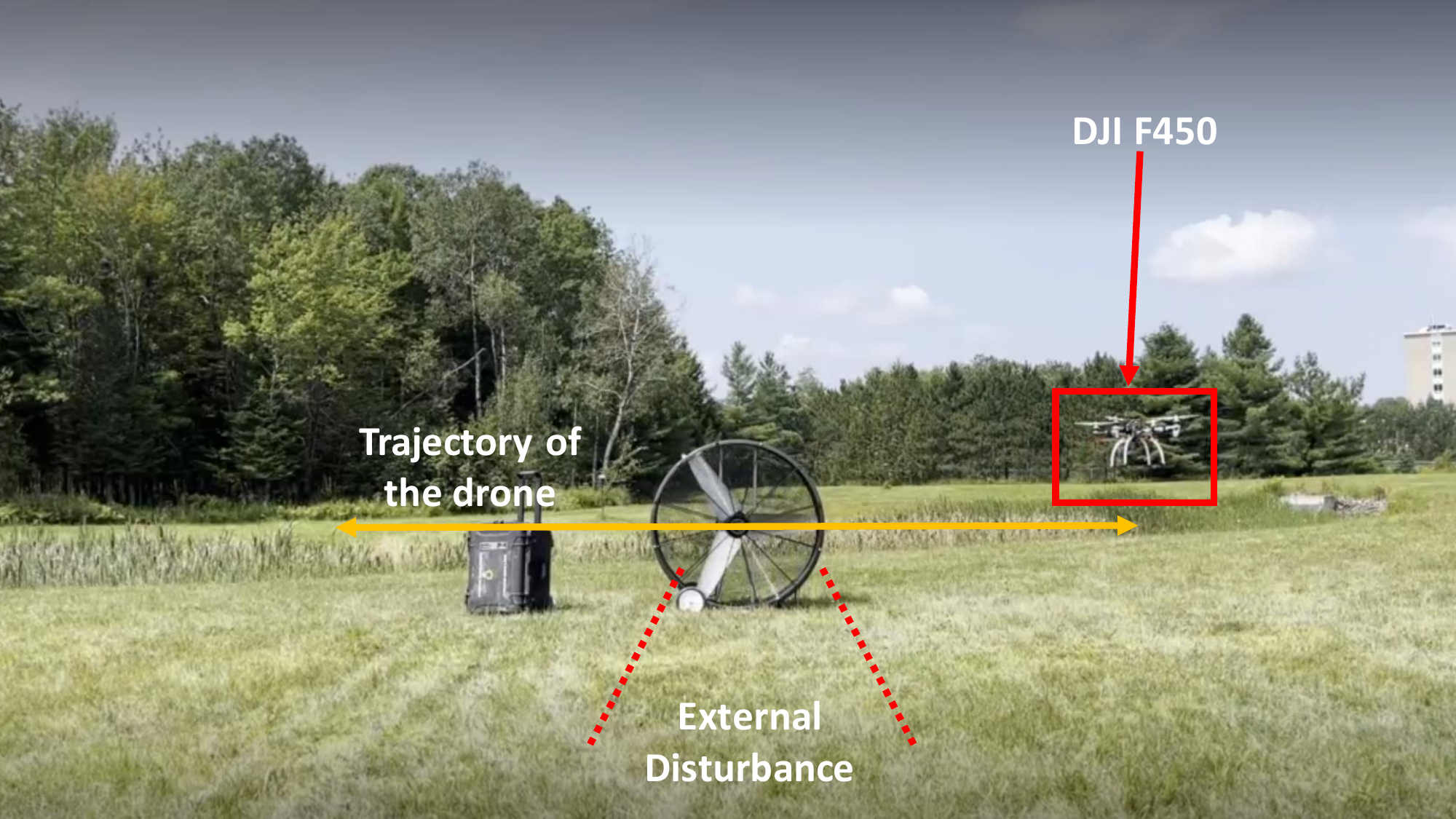}
    \caption{Experimental setup for testing the impact of external disturbance. }
    \label{fig:experimental_setup}
\end{figure}

Central to our experimental setup, shown in Fig. \ref{fig:experimental_setup}, was a 3-foot fan used to emulate consistent wind gusts of 4 m/s. To examine the drone's stability, its response rate and its overall performance under induced wind conditions, we systematically maneuvered the drone across the wind gusts produced by the fan.. 

Furthermore, we conducted tests using various payloads, ranging from 0 to 0.95 kg. This was essential for examining the effects of different thrust-to-weight ratios (T/W), which are known to drastically impact a drone's capacity to withstand external disturbances.
As a general guideline, to achieve stable flight in high wind conditions, the T/W should exceed 3. For mild wind conditions, a ratio between 2 and 3 is sufficient, while for indoor flights and slow-speed maneuvers, a ratio around 1.5 is the lowest advisable \cite{HalfChromeDrone}. We evaluated the correlation between T/W adjustments and motor commands for the F450, showcasing a shift from an average motor command of 46\% (2:1) to 74\% (1.25:1). 


The main goal of the developed algorithm is to assess real time risk related to external disturbance. Moreover, we aim to predict if the wind is going to be too dangerous in the near future. 
Our approach was compared to ArduCopter's embedded wind estimator \cite{ArduPilot} to gauge the wind. For this estimation to be effective, it must be both precise and quick to respond. A swift response is essential as it helps detect brief gusts, which pose significant challenges for drone operations. 

\begin{figure}[tbh]
    \centering
    \includegraphics[width=0.5\textwidth]{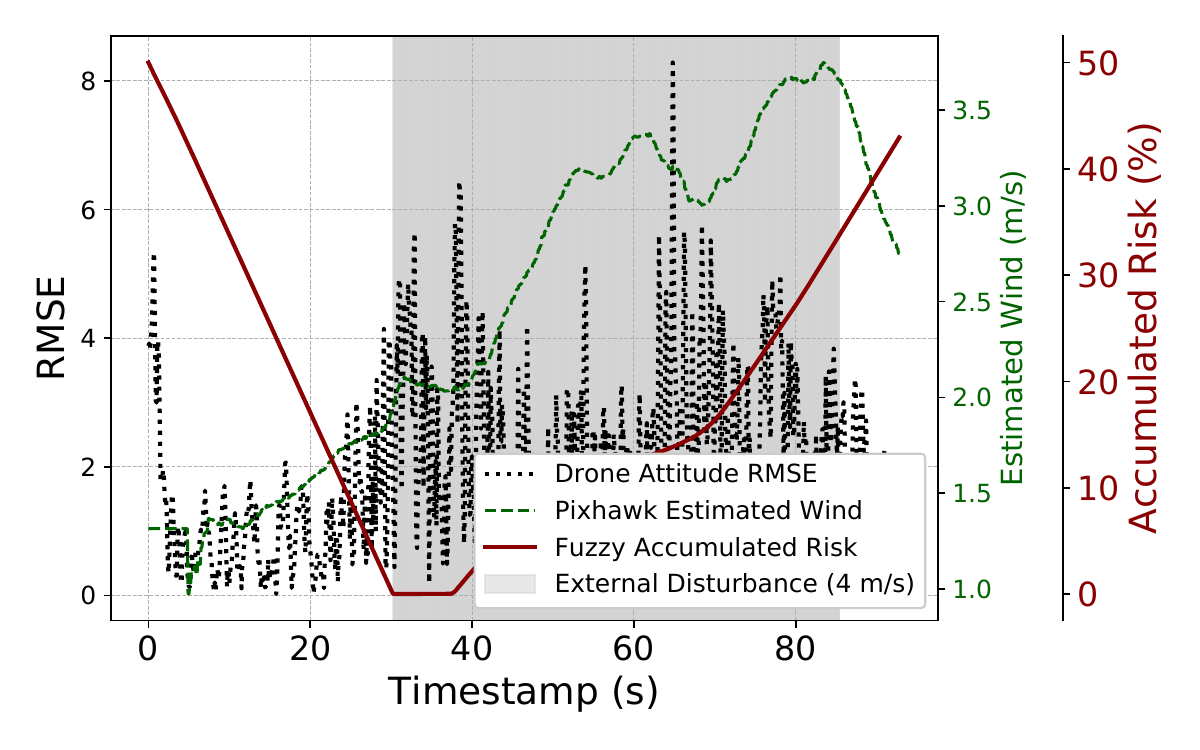}
    \caption{
    Comparison between Pixhawk wind estimation (black) and the proposed approach (red) in a steady 4 m/s wind (from 30s to 83s).
    }
    \label{fig:flight_steady}
\end{figure}

\begin{figure}[tbh]
    \centering
    {\includegraphics[width=0.5\textwidth]{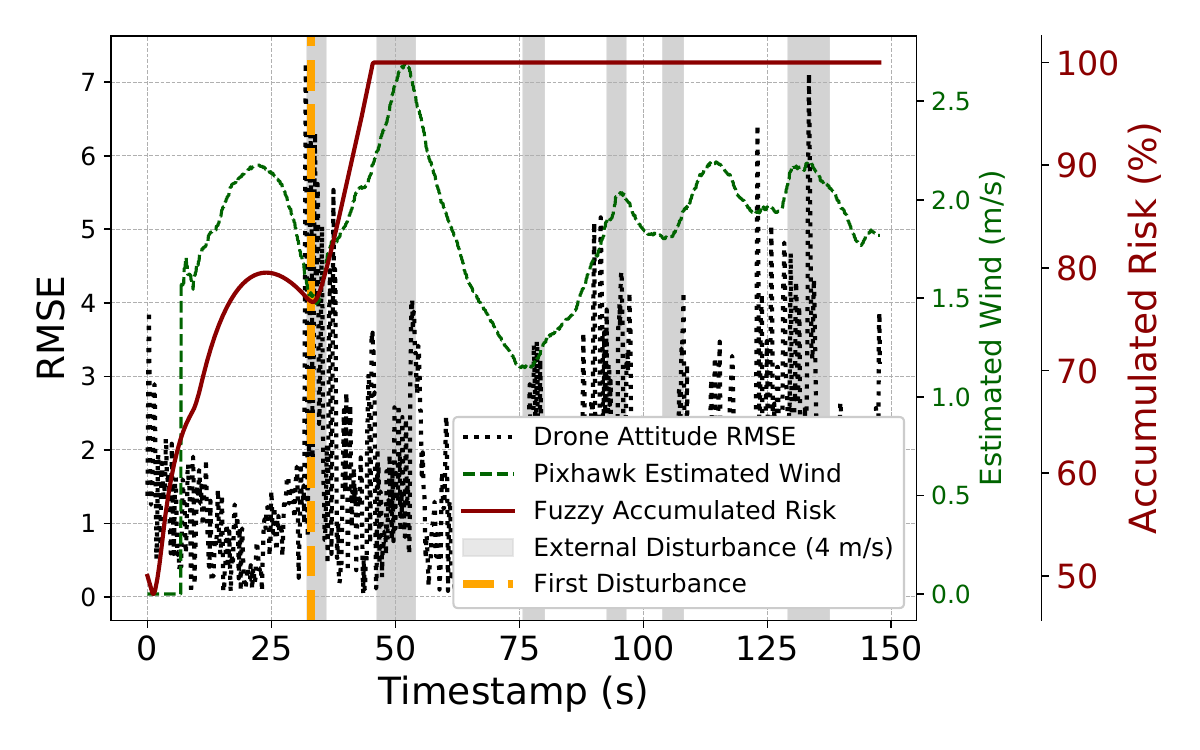}}
    \caption{Comparison between Pixhawk wind estimation (black) and the proposed approach (red) in a intermittent 4 m/s wind.}
    \label{fig:flight_perturbation}
\end{figure}

In Fig. \ref{fig:flight_steady}, the drone was subjected to a consistent 4 m/s wind, generated by the fan, from t=30s to 83s. The ArduCopter estimator took approximately 20 seconds to stabilize. Moreover, our method accurately assessed a minimal risk level for the initial 30 seconds when the drone was unaffected by any disturbances. In Fig. \ref{fig:flight_perturbation}, the ArduCopter's estimation did not align with the perturbation. Yet, our suggested methodology flagged the flight as hazardous (with a risk index surpassing 75\%) the moment the drone encountered its first disturbance at around 33s. From the test results, it's evident that the proposed method surpasses ArduCopter's wind estimation capabilities and excels at identifying elevated risk levels due to various external disturbances.

\begin{figure}[htbp]
    \centering
    \includegraphics[width=0.5\textwidth]{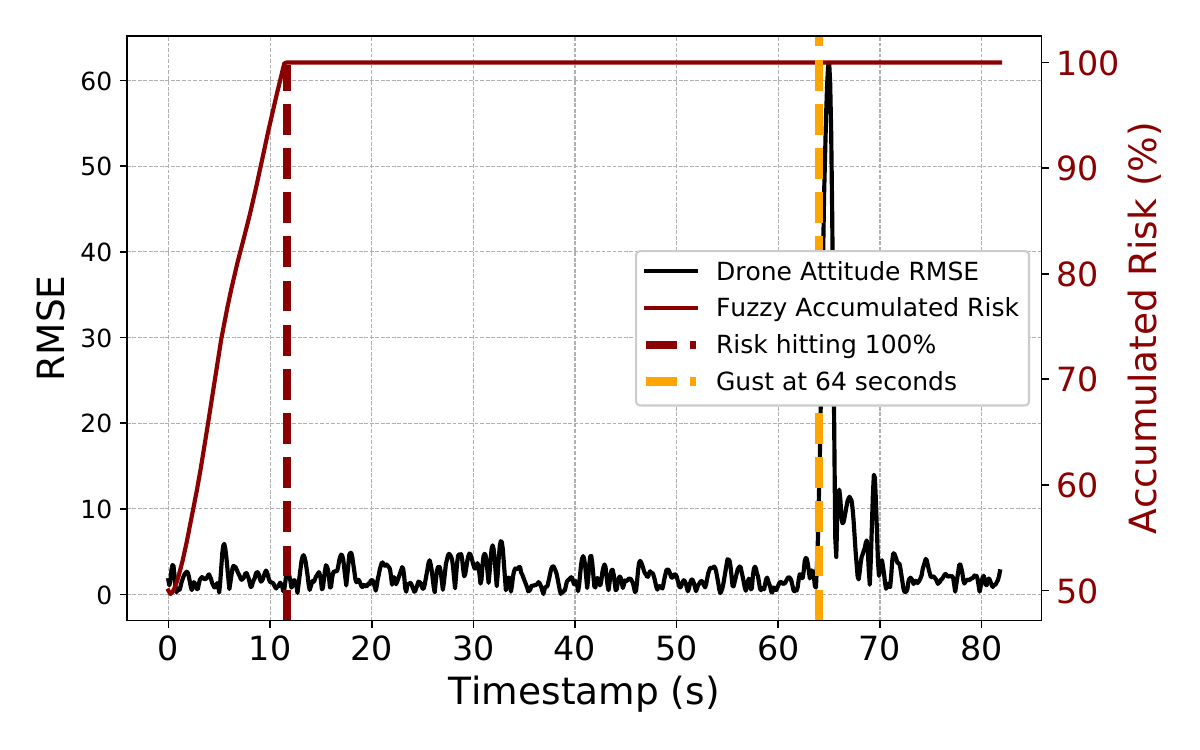}
    \caption{Risk estimation on a dangerous flight with NADILE.}
    \label{fig:Nadile_flight}
\end{figure}

To confirm the generality of our method, we tested our algorithm on a log of a dangerous flight with the NADILE. The pilot deemed the wind conditions to be safe enough to fly, but an unexpected gust almost led to the drone crashing. Fig. \ref{fig:Nadile_flight} demonstrates that our method would have identified the flight as risky 60s prior to the gust.

\section{Conclusion}
In this study, we proposed a new approach to assessing risk posed by external disturbances by using standard motor commands. The method does not need any additionnal sensors, drone models or datasets. This makes it a versatile approach for various multirotors. We showcased its effectiveness through real-world tests and compared it against the well-known ArduCopter wind estimator. The algorithm could be integrated into an onboard computer to either relay the risk level to the operator or allow an autonomous system to act based on the perceived risk, potentially enhancing flight capabilities. In future studies, the algorithm's performance could be enhanced by incorporating additional inputs into the risk estimation, including battery level, sensor malfunctions, and communication degradation.

\bibliographystyle{IEEEtran}
\bibliography{reference}

\end{document}